\documentclass[letterpaper, 10 pt, conference]{ieeeconf}  
\usepackage{soul,color}
\usepackage{todonotes}
\usepackage{amsmath,bm}
\usepackage{subcaption}
\usepackage{float}
\usepackage{cite}
\usepackage{multirow}

\IEEEoverridecommandlockouts                              

\title{\LARGE \bf
Combining Time-Dependent Force Perturbations in Robot-Assisted Surgery Training}

\author{Yarden Sharon$^{1}$, Daniel Naftalovich$^{2}$, Lidor Bahar$^{1}$, Yael Refaely$^{3}$, and Ilana Nisky$^{1}$
\thanks{*This work was supported in part by the Helmsley
Charitable Trust through the Agricultural, Biological and
Cognitive Robotics Initiative and by the Marcus Endowment
Fund both at Ben-Gurion University of the Negev
and the ISF Grant No. 327/20. YS was supported by the Besor scholarship and the Israeli Planning and Budgeting Committee scholarship.}
\thanks{$^{1}$YS, LB, and IN are with the Department of Biomedical Engineering and Zlotowski Center for Neuroscience, Ben-Gurion University of the Negev, Israel.
        {\tt\small shayar@post.bgu.ac.il, lidorba@post.bgu.ac.il, nisky@bgu.ac.il}}%
\thanks{$^{2}$DN is with the Department of Computational \& Mathematical Sciences, California Institute of Technology, USA, and 
Keck School of Medicine of USC, University of Southern California, USA.
        {\tt\small nafty@caltech.edu }}%
    \thanks{$^{3}$YR is with the Thoracic Surgery Unit, Soroka Medical Center, Israel.
{\tt\small yaelrefaely@clalit.org.il}}%

}

\begin{document}

\maketitle
\thispagestyle{empty}
\pagestyle{empty}

\begin{abstract}
Teleoperated robot-assisted minimally-invasive surgery (RAMIS) offers many advantages over open surgery. However, there are still no guidelines for training skills in RAMIS. Motor learning theories have the potential to improve the design of RAMIS training but they are based on simple movements that do not resemble the complex movements required in surgery. To fill this gap, we designed an experiment to investigate the effect of time-dependent force perturbations on the learning of a pattern-cutting surgical task. Thirty participants took part in the experiment: (1) a control group that trained without perturbations, and (2) a 1Hz group that trained with 1Hz periodic force perturbations that pushed each participant's hand inwards and outwards in the radial direction. We monitored their learning using four objective metrics and found that participants in the 1Hz group learned how to overcome the perturbations and improved their performances during training without impairing their performances after the perturbations were removed. Our results present an important step toward understanding the effect of adding perturbations to RAMIS training protocols and improving RAMIS training for the benefit of surgeons and patients.
\end{abstract}

\section{INTRODUCTION}
In robot-assisted minimally-invasive surgery (RAMIS), surgeons use robotic manipulators to control the movements of instruments, which are inserted into patients' bodies via small incisions \cite{maesoEfficacyVinciSurgical2010}. RAMIS offers many advantages over open surgery \cite{moorthyDexterityEnhancementRobotic2004}. However, to reap the benefits of RAMIS, surgeons must be well trained to use the robotic systems \cite{crawfordEvolutionLiteratureReview2018}. Currently, there are training guidelines for open and laparoscopic surgery, but not for RAMIS \cite{ahmedDevelopmentStandardisedTraining2015}. 

An important step to mitigate this gap is to study how surgeons acquire RAMIS skills, and to discover what affects their learning. Previous studies attempted to influence the acquisition of RAMIS skills, e.g. by choosing which exercises the trainees perform. For example, in \cite{gurungAcceleratedSkillsAcquisition2020} learning was sped up by focusing the training on high difficulty exercises. In \cite{marianiSkillOrientedPerformanceDrivenAdaptive2021}, an adaptive training protocol was proposed in which the exercises are chosen based on the trainee's performance, leading to improved performance  of participants compared to participants who chose their exercises on their own.

Another way to affect RAMIS skill acquisition is to combine task execution examples of an experienced user during training. For example, in \cite{yangEffectivenessIntegratedVideo2017} participants who watched their own performances along with a video of an expert performing the same exercise learned better than participants who did not have access to such feedback. In addition, many research groups developed training platforms that combine haptic guidance that enables the trainee to follow the movement of an experienced user who performs the exercise \cite{jacobsImpactHapticLearning2007,shahbaziMultimodalSensorimotorIntegration2018,abdelaalPlayMeBack2019}.

Important sources of knowledge about how to train RAMIS surgeons are motor learning theories \cite{jarcRobotassistedSurgeryEmerging2015}. Motor learning studies investigate the different processes that enable learning. These studies define skill acquisition as an improvement in performance beyond previous levels or the acquisition of completely novel abilities \cite{krakauerHumanSensorimotorLearning2011}. Adaptation is defined as when participants improve their performances in response to altered conditions such as a perturbing force field \cite{shadmehrAdaptiveRepresentationDynamics1994} or visuomotor transformations \cite{krakauerLearningVisuomotorTransformations2000}. It is important to note that in contrast to skill acquisition, in adaptation, the participants can restore their baseline performance but will not improve
beyond it \cite{krakauerHumanSensorimotorLearning2011}.

Recent findings in motor learning propose ways to affect learning. For example, in reach movements, manipulating the consistency
of the perturbations \cite{castroEnvironmentalConsistencyDetermines2014,wuTemporalStructureMotor2014,duarteEffectsRoboticallyModulating2015}, the error that is presented to the trainee \cite{herzfeldMemoryErrorsSensorimotor2014,vanderkooijVisuomotorAdaptationHow2015,diederenScalingPredictionErrors2015}, and the balance between punishment and reward \cite{galeaDissociableEffectsPunishment2015}, can affect the rate of adaptation. In error-based learning, the sensorimotor system is hypothesized to estimate the error between the desired or predicted outcome of a movement and the actual outcome, and update the motor commands in the following movement \cite{wolpertPrinciplesSensorimotorLearning2011}. These ideas may be used to optimize surgical
skill learning \cite{jarcRobotassistedSurgeryEmerging2015}, but there is a gap between the current knowledge that is based on simple movements and the needed knowledge to train RAMIS surgeons, who perform complex motor tasks. There are a few studies that begin to fill the gap \cite{m.m.coadTrainingDivergentConvergent2017,oquendoRobotAssistedSurgicalTraining2019}, but more work is needed to develop efficient training protocols. 

In this study, for the first time, we examine the effect of time-dependent perturbations on the learning of a surgical task. In our experiment, the participants cut circles drawn on gauze while they were exposed to perturbations that alternatingly pushed their hands inwards and outwards in the radial direction. We chose a time-dependent perturbation because while acting on the human body, surgeons encounter various time-dependent perturbations; for example, as a result of the periodic movement of the heart and blood vessels. To develop RAMIS training protocols, we need to understand how surgeons learn to deal with such perturbations, and their effects on performance.

More specifically, it is important to understand whether surgeons can improve their performance under time-dependent perturbations. Our task has a clear desired path, and the perturbations increase the error between the desired movement and the actual movement. Hence, we hypothesized that the motor system would adjust motor commands and reduce error with training. If surgeons do manage to improve performances during exposure to the perturbations, it is important to test whether this learning impairs their ability to cope with other conditions (such as an environment without perturbations), and whether it gives them resistance to other perturbations as well. 
We designed our experiment to answer four specific research questions:
\begin{itemize}
\item \textbf{Q1} -- Whether participants that are exposed to force perturbations can learn and improve their performance under these perturbations?
\item \textbf{Q2} -- Whether training with force perturbations impairs the performance when the perturbations are removed?
\item \textbf{Q3} -- Whether training with force perturbations can give an advantage when later encountering these perturbations, compared to those who trained without perturbations?
\item \textbf{Q4} -- Whether training with force perturbations can give an advantage when encountering different types of perturbations, compared to those who trained without perturbations?
 \end{itemize}

A preliminary version of this study which included an initial analysis of the pilot experiment was presented in an extended abstract form \cite{sharonPreliminaryAnalysisLearning2020}.

\section{METHODS}
    
    \subsection{Experimental Setup} 
    \subsubsection{the da Vinci Research Kit (dVRK)}
\begin{figure}[t]
			\centering
            \includegraphics[width=\columnwidth]{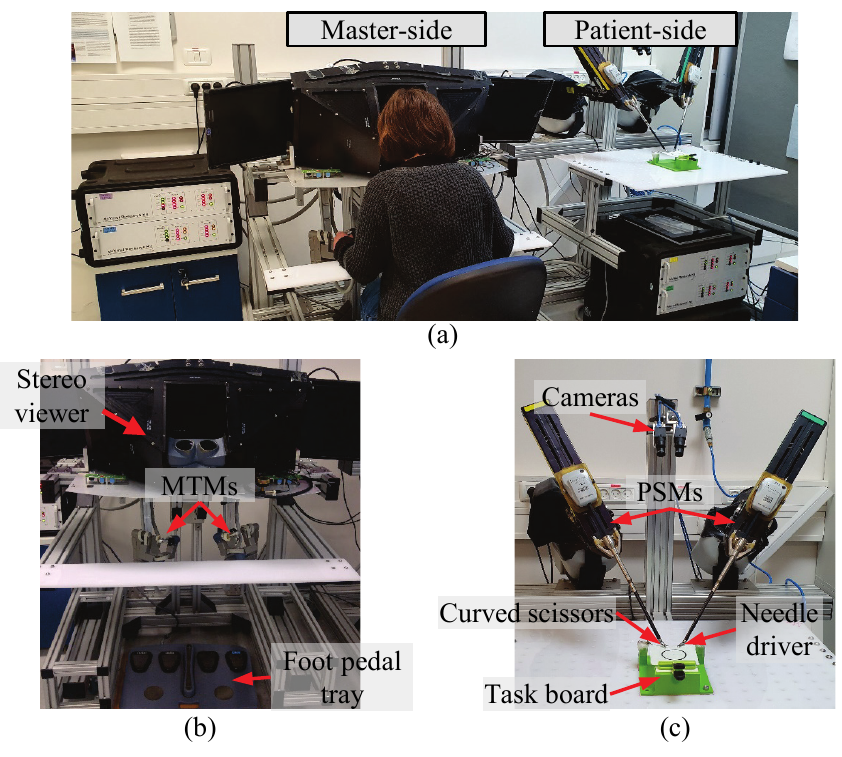}
			\caption{The dVRK. (a) The participant sits in the master-side and uses the MTMs to remotely teleoperate the PSMs in the patient-side. (b) The master-side. (c) The patient-side.}
			\label{fig:dVRK}
		\end{figure}

        The dVRK is a development platform for researchers in the field of RAMIS \cite{kazanzidesfOpensourceResearchKit2014}, provided by Intuitive Surgical. Its hardware consists of components from the first-generation da Vinci Surgical System \cite{guthartIntuitiveTextsuperscriptTMTelesurgery2000}. Our dVRK (Fig. \ref{fig:dVRK}) consists of a pair of Master Tool Manipulators (MTMs), a pair of Patient Side Manipulators (PSMs), a foot-pedal tray, a high resolution stereo viewer, and four manipulator interface boards. In this study, the participants sat at the master-side (Fig.\ref{fig:dVRK}.b), and used the MTMs to remotely teleoperate curved scissors and a large needle driver at the patient side, where the task board was placed (Fig. \ref{fig:dVRK}.c). The MTMs and PSMs electronics were connected via firewire to a single Ubuntu (UNIX) OS computer with an Intel Xeon E5-2630 v3 processor. The vision system consisted of a pair of Blackfly S cameras (FLIR Integrated Imaging Solutions Inc.) that acquired the visual scene in the patient side. The cameras were fixed such that the task board was in the center of the field of view. The participants could not control the cameras or the zoom. The video was broadcasted to the stereo viewer, which presented a 3D view at 35 Hz refresh rate and with 1080 X 810 resolution per eye. The visual information was transmitted using a custom-developed software on a dedicated computer. The movement scaling was set to 0.4, such that each movement of the PSM was 0.4 times of the movement of the MTM, and the participants could reach the entire workspace without using the clutch. 
        
    \subsubsection{The Pattern-Cutting Task}
\begin{figure}[t]
			\centering
            \includegraphics[width=0.95\columnwidth]{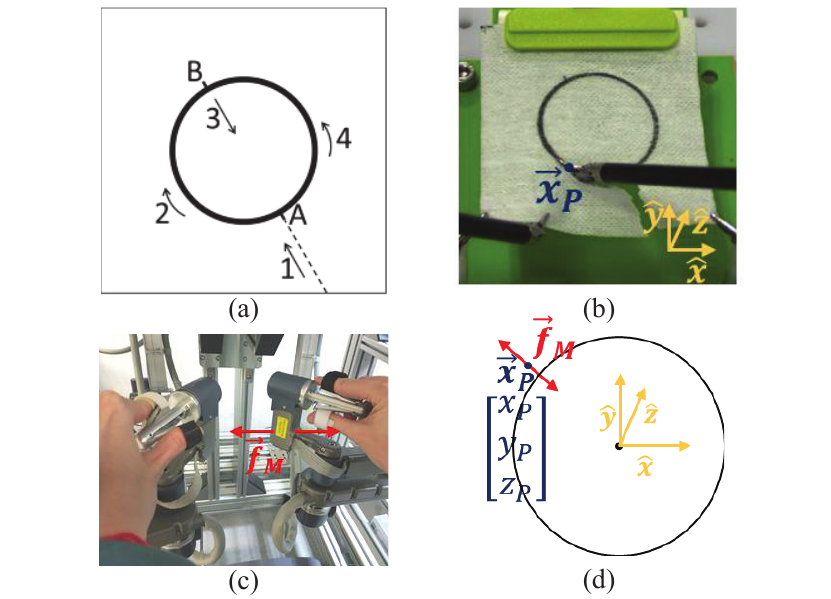}
			\caption{The pattern-cutting task and the force perturbations. (a) The task sequence. (b) The task board and the two PSM tools (right – curved scissors, left – large needle driver). (c) The MTMs. (d) The circle; the position of the right PSM's tip $\vec{\bm{x}}_P$ (blue); the force applied on the participant's hand $\vec{\bm{f}}_M$ (red), and the reference base frame (yellow).}
			\label{fig:TheTask}
		\end{figure}

        The task in this experiment was based on the FLS pattern-cutting task \cite{FLSManualSkills2014}, and modified to our needs. In this task, participants used their right hands to control curved scissors and cut a 5cm diameter circle drawn on a two-layered 10X10cm non-woven gauze. The width of the black circle line was 2mm. To complete the task participants could cut one layer or both layers. The task sequence consisted of (Fig. \ref{fig:TheTask}.a): (1) cutting the gauze toward point A; (2) cutting along the left half of the circle, until point B; (3) moving the scissors back to point A; (4) cutting the right side of the circle. The participants used their left hands to control a large needle driver and maintain the tension of the gauze. Participants were instructed to cut as quickly and as accurately as possible, which meant within the black line.
        
    \subsubsection{Force Perturbations}
  In some of the trials during the experiment, planar radial force perturbations were applied on each participant’s hand using the right MTM -- away from the center of the circle and toward the center, alternatingly (Fig. \ref{fig:TheTask}(c)-(d)):
    \begin{equation}
    \label{eq:perturbations}
        \overrightarrow{\boldsymbol{f}}_{M}=A\left[\frac{x_{P}}{\sqrt{x_{P}^{2}+y_{P}^{2}}}, \frac{y_{P}}{\sqrt{x_{P}^{2}+y_{P}^{2}}}, 0\right]^T,
    \end{equation}
    where $x_P$ and $y_P$ are the x and y position coordinates of the scissors' tip, relative to the center of the circle (see Fig. \ref{fig:TheTask}.d), and A defines the type of the trial. There were three types of trials during the experiment (see Video 1):
    \begin{itemize}
       \item No perturbations: $A =0$.
       \item 1Hz perturbations: $A = sin(2\pi t)$.
       We chose the 1N maximal force applied in this type of trial such that it was noticeable but still allowed task completion. 
       \item Unpredictable perturbations: $A=\frac{\sum_{1}^{5} \sin (2 \pi f t)}{\sqrt{5}} ; f \sim U[0.3 H z, 1 H z]$. The force was a combination of five sine waves with different frequencies, known to be unpredictable to human users  \cite{avrahamPerceivingRobotsHumans2012}. We normalized the amplitude by dividing the sum by $\sqrt{5}$, leading to power that was equal to that of the 1Hz perturbations.
    \end{itemize}
    
    To make sure the perturbations worked in the radial directions, the gauze was placed so that the center of the circle was placed on the origin of the reference base frame (Fig. \ref{fig:TheTask}.d). Before each trial, the PSM tool pointed to the origin of the reference base frame and the experimenter placed the marked center of the circle in the right position. 

    \subsection{Experimental Protocol}
    Thirty right-handed volunteers without surgical background (aged 21-30; 15 females) participated in the experiment after signing an informed consent approved by the Human Participants Research Committee of Ben-Gurion University of the Negev. The participants were instructed on how to use the dVRK and how to perform the pattern-cutting task. Since the perturbations are based on the position of the circle, the experimenter emphasised the importance of keeping the gauze in its place, i.e., not pulling it in a way that distorts the marked circle. The instructions were accompanied by videos of the pattern cutting task -- a task that performed according to the instructions, and a task in which the participant stretches the gauze in a prohibited way, and also cuts with many errors (not within the black line). Before the experiment, each participant practiced the following actions with the robot: pressing the right pedal to start teleoperation, moving the right tool through the task sequence (without cutting), catching the gauze with the left tool, and cutting a straight line with the right tool.  
    
\begin{figure}[t]
			\centering
            \includegraphics[width=0.8\columnwidth]{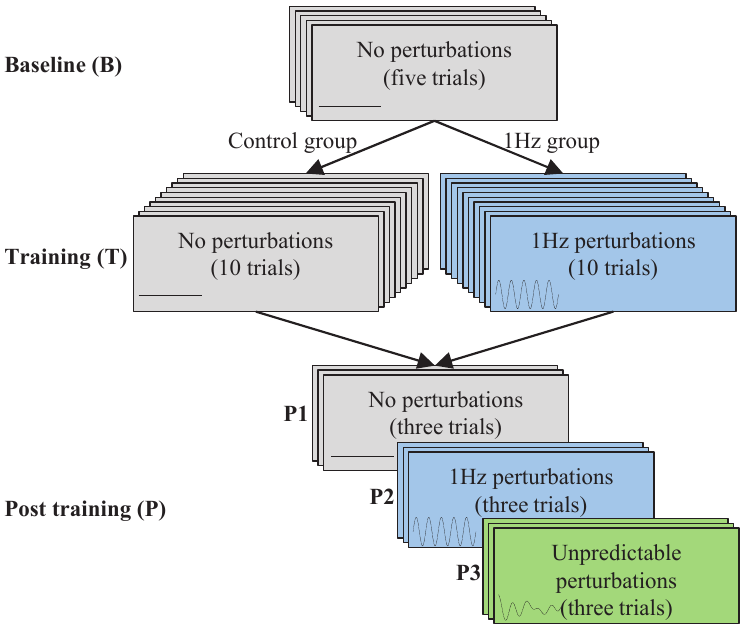}
			\caption{The experimental protocol.}
			\label{fig:Protocol}
		\end{figure}

    The participants were randomly assigned into two groups (15 participants per group): a 1Hz group, and a control group. During the experiment each participant performed 24 consecutive trials (Fig. \ref{fig:Protocol}):
        \begin{itemize}
        \item \textbf{Baseline (B)} -- five trials without perturbations. These trials were the same for both groups, and their purpose was to assess baseline performances of each participant. 
        \item \textbf{Training (T)} -- 10 trials that were used to answer Q1: the control group trained without perturbations, and the 1Hz group trained with 1Hz perturbations.
        \item \textbf{Post training (P)} -- nine trials for testing the effect of the training on performance post training, i.e. answering  Q2-Q4; these trials were the same for both groups. The first three trials \textbf{(P1)} were without perturbations, and were included to assess the performances of both groups after the different training protocols (Q2). The next three trials \textbf{(P2)} were with 1Hz perturbations, and were included to asses the resistance to 1Hz perturbations (Q3). The last three trials \textbf{(P3)} were with unpredictable perturbations, and were included to asses the resistance to new perturbations, which participants had not practiced before (Q4). The five frequencies of the unpredictable trials were different between the three trials, but the same for all the participants. 
        \end{itemize}
        
\subsection{Data Acquisition and Segmentation}
 All the kinematic data of the MTMs and PSMs were recorded at 100Hz. The video data of the left camera were recorded at 35Hz. In addition, all the cut circles were scanned. In 20 trials out of the 720 trials there were technical issues that resulted in the trial stopping before the task completion. In all those trials, participants completed the trial after fixing the issue, and these trials were not included in the analysis. We used the recorded videos and the scissors' opening angle to manually segment each trial into its stages.
\subsection{Metrics}
We used four metrics to assess the progress of the participants: \textit{combined error-time}, \textit{path length}, \textit{number of cuts}, and \textit{perturbation MSE}. The first two metrics quantified the task performance, and the other two metrics allowed us to follow different approaches of the participants. 

\subsubsection{Combined error-time}
participants were instructed to cut as quickly and as accurately as possible. According to the speed-accuracy trade-off \cite{fittsInformationCapacityHuman1954}, we expected that shorter task times might be accompanied by larger errors and vice versa. Hence, we combined these two measures of performance. 
The task time was calculated as:
\begin{equation}
    TT = t_{end}-t_{start},
\end{equation}
where $t_{start}$ is the time when the participant first closes one of the tools on the gauze, and  $t_{end}$ is the time of the last cut, when the circle was completely removed from the gauze.

\begin{figure}[b]
			\centering
            \includegraphics[width=\columnwidth]{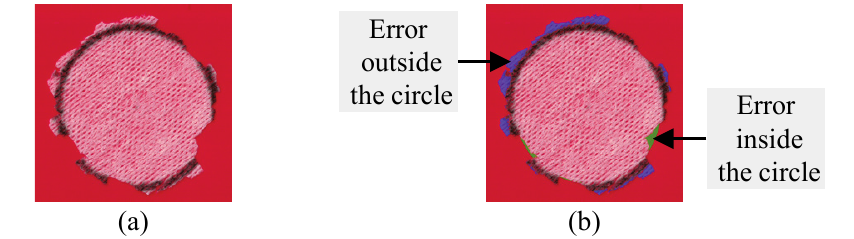}
			\caption{Example of total error calculation. (a) The cut circle. (b) The circle with marked error areas.}
			\label{fig:TotalError}
		\end{figure}

To quantify the amount of errors, we used a custom-written image processing algorithm (MATLAB) that detected error areas in the scanned circles. The error areas were defined as areas in which the cutting was not on the line -- outside or inside the circle (Fig. \ref{fig:TotalError}). The total error was calculated as:
 \begin{equation}
    TE = E_{outside}+E_{inside},
\end{equation}
where $E_{outside}$ and $E_{inside}$ are the numbers of pixels in the error areas outside and inside the circle, respectively.  

The combined error-time of trial $j$ was calculated as \cite{liesefeldCombiningSpeedAccuracy2019}:
 \begin{equation}
    CET_j = \frac{TT_j-\overline{TT}}{S_{TT}}+\frac{TE_j-\overline{TE}}{S_{TE}},
\end{equation}
where $\overline{TT}$ and $\overline{TE}$ are the average values of the task times and the total errors across all the trials of all the participants in the experiment, $S_{TT}$ and $S_{TE}$ are the standard deviations. Note that the value of this metric can be positive or negative and that lower value means better performance.

\subsubsection{Path Length}
this metric is a common measure of surgical skill performance which quantifies the economy of motion, and it was calculated as: 
\begin{equation}
PL = \sum_{i=1}^{N-1} ||\vec{\bm{x}}_P[i+1]-\vec{\bm{x}}_P[i]||_2,
\end{equation}
where $\vec{\bm{x}}_P[i]$ is the position of the scissors' tip at the i\textsuperscript{th} sample, and N is the number of samples.

\subsubsection{Number of cuts}
one approach to dealing with the perturbations may be to time the cuts according to the perturbations. Such an approach would lead to a change in the number of cuts during training. Hence, we counted the number of times the scissors were closed during the task. To find these closing events, we used the MATLAB function \texttt{findpeaks()} to find the local minima of the recorded scissors' opening angle. We included in the analysis only cuts that were made for cutting the circle itself (i.e, steps 2 and 4, Fig. \ref{fig:TheTask}.a).

\subsubsection{Perturbation MSE}
\begin{figure}[t]
			\centering
            \includegraphics[width=0.8\columnwidth]{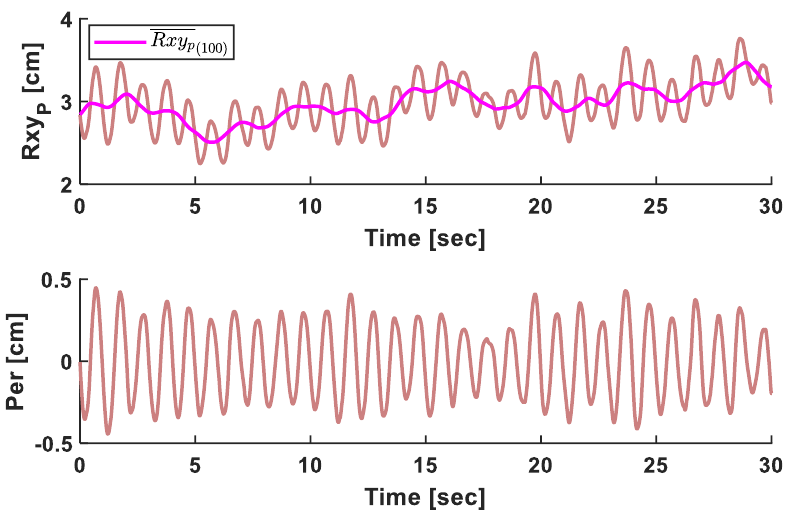}
			\caption{Isolating the perturbation from the recorded path.}
			\label{fig:Perturbation}
		\end{figure}

this metric quantifies the effect of the perturbations on the participant's movement. The higher the value of this metric, the greater the 1Hz radial fluctuations in the hand movement are. The size of the fluctuations is a measure of coping with the perturbations by either adaptation---i.e., actively cancelling the perturbation be opposing forces---or by increasing of arm stiffness.
To calculate this metric, we first extracted the planar radial component of the scissors' path (Fig. \ref{fig:Perturbation}, upper panel):
\begin{equation}
Rxy_p[i] = \sqrt{{x_P[i]}^2+{y_P[i]}^2}.
\end{equation}
Where $x_P$ and $y_P$ are the x and y position coordinates of the scissors' tip.
We then filtered $Rxy_p$ using a moving average of 100 samples:
\begin{equation}
\overline{Rxy_p}_{(100)}[i] =
\frac{1}{100} \sum_{k=-50}^{49} Rxy_p[i+k].
\end{equation}
At the edges of the signal, the average of the available samples was calculated. The 1Hz perturbation was periodic with a period duration of 1sec, and there were 100 samples per second. Therefore, $\overline{Rxy_p}_{(100)}$ is the radial component without the contribution of the perturbation  (Fig.\ref{fig:Perturbation}, magenta line). To isolate the contribution of the perturbation (Fig. \ref{fig:Perturbation}, lower panel) we used :  
\begin{equation}
Per[i] = Rxy_p[i]-\overline{Rxy_p}_{(100)}[i] 
\end{equation}
The \textit{perturbation MSE} was calculated as:
\begin{equation}
PerMSE = \frac{1}{N}\sum_{i=1}^{N}  (Per[i])^2
\end{equation}
We calculated $PerMSE$ for steps 2 and 4 -- cutting the circle (Fig. \ref{fig:TheTask}.a), and summed them to get one metric value per trial. Since the unpredictable perturbations in P3 (Fig. \ref{fig:Protocol}) was a combination of five sine waves with different frequencies, we did not calculate this metric on trials in P3.

\subsection{Statistical analysis}
Our metrics were not normally distributed, and therefore we tested our hypotheses using permutation tests \cite{goodPermutationTestsPractical2013}. For each participant we calculated the average metric values of: the last three trials of the baseline ($B$); the first three trials of the training ($T_{start}$); the last three trials of the training ($T_{end}$); the post training trials without perturbations ($P1$); the post training trials with 1Hz perturbations ($P2$); the post training trials with unpredictable perturbations ($P3$).

 We first tested Q1 -- whether participants that are exposed to force perturbations can learn and improve their performance under the perturbations. For each metric, we calculated the difference between $T_{start}$ and $T_{end}$ of each participant. We then used a matched-pair permutation test for each group to test whether the mean of these differences was significantly different than zero. Additionally, we used a permutation test to test whether the means of the individual differences between $T_{start}$ and $T_{end}$ were significantly different between the two groups (control and 1Hz). 

We then tested Q2 -- whether training with force perturbations can impair the performance when the perturbations are removed. We performed a permutation test to test whether there was a significant difference between the mean $P1$ values of the two groups. In addition, we calculated the improvement between the baseline trials and P1 trials ($P1-B$) and used them for another permutation test between the two groups. 

Next, we tested Q3 -- whether training with force perturbations can give an advantage when encountering these perturbations, compared to those who trained without perturbations. We performed two permutation test between the  groups: (1) $P2$ values, (2) $P2-P1$ values.

Lastly, we tested Q4 -- whether training with force perturbations can give an advantage when encountering different perturbations, compared to those who trained without perturbations. We performed two permutation test between the  groups: (1) $P3$ values, (2) $P3-P1$ values.

To control for multiple comparisons, we used the Bonferronni correction. For each of the research questions separately, we multiplied the p-values by the number of tests (two or three). Statistical significance was determined at the 0.05 threshold.

\section{Results}
\begin{figure}[t]
	\centering
    \includegraphics[width=\the\columnwidth]{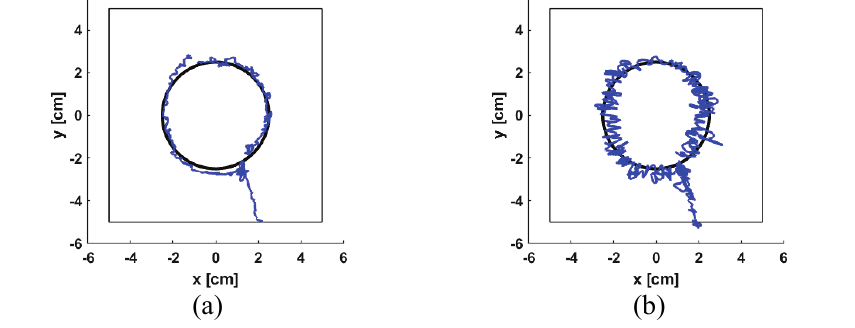}
	\caption{Examples of the recorded path of the scissors. (a) The last baseline trial -- without perturbations, and (b) the same participant’s first training trial with 1Hz perturbations.}
	\label{fig:CutExamples}
\end{figure}

Fig. \ref{fig:CutExamples} shows the scissors’ path when the participant cut the circle. The deviations from the circle in Fig. \ref{fig:CutExamples}(b) are more prominent than those in Fig. \ref{fig:CutExamples}(a), showing that the perturbations affect the scissors’ path. Fig. \ref{fig:Results1} depicts the values of the four metrics during the experiment trials. Fig. \ref{fig:Results2} presents the average values for the different experiment stages (left panel), and the average values of the individuals differences between the stages (right panel). 
Videos 2-5 present the videos of the trials with the lowest and highest scores for each metric.

For all the four metrics, there was no significant difference between the two groups at the baseline stage (p\textsubscript{combined error-time} = 0.2165,
p\textsubscript{path length} = 0.9985, 
p\textsubscript{number of cuts} = 0.1810, 
p\textsubscript{perturbation MSE} = 0.7355). In the following subsections we will describe the results of the tests we conducted to examine our research questions. Table \ref{tab:Results} summarizes the statistical analysis.

\subsection{Q1 -- The effect of the training protocols on the learning curves and approaches}
\begin{figure}[!t]
	\centering
    \includegraphics[height=15.5cm]{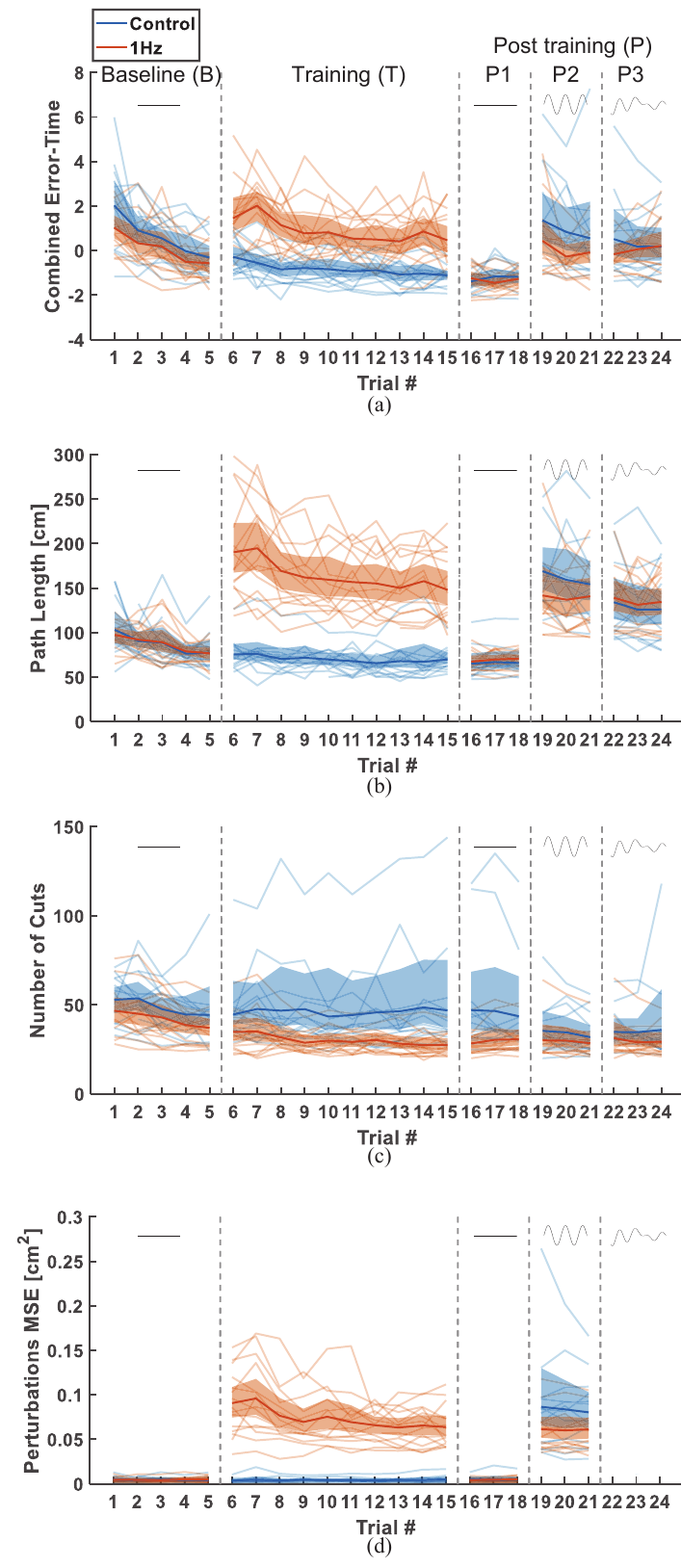}
	\caption{The values of the four metrics as a function of trial number. Pale color lines are individual scores, dark color lines are means, and shaded areas are 95\% bootstrap confidence intervals.}
 	\label{fig:Results1}
\end{figure}

\begin{figure}[!t]
	\centering
    \includegraphics[height=15.5cm]{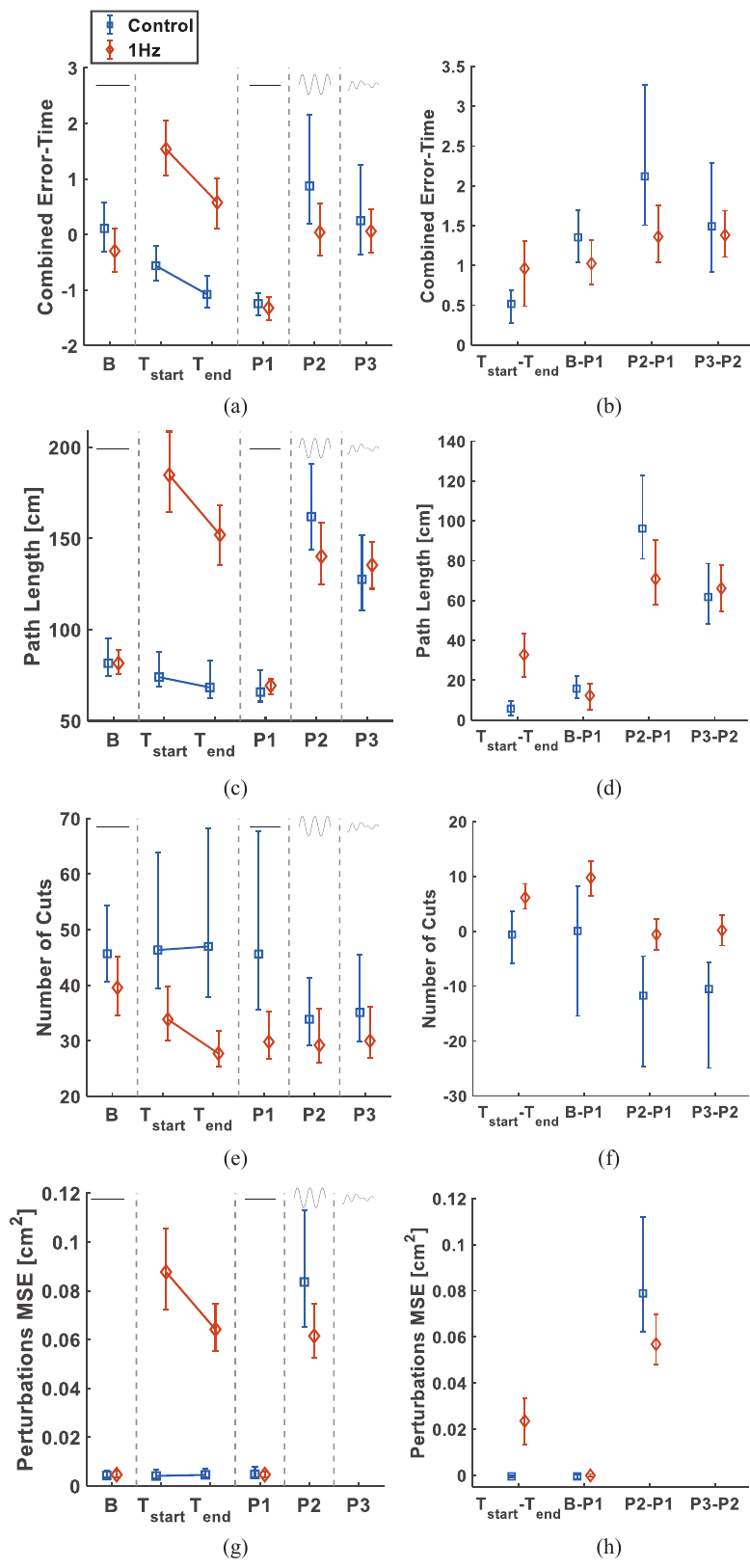}
	\caption{The left panel presents the average values of the four metrics in the different stages of the experiment: last three baseline trials (B), first three training trials (T\textsubscript{start}), last three training trials (T\textsubscript{end}), post training trials without perturbations (P1), post training trials with 1Hz perturbations (P2), and post training trials with unpredictable perturbations (P3). The right panel present the average values of the individual differences between the stages: T\textsubscript{start}-T\textsubscript{end}, B-P1, P2-P1, P3-P2. Marker are means, error bars are 95\% bootstrap confidence intervals.}
 	\label{fig:Results2}
\end{figure}

Participants from both groups improved their \textit{combined error-time} and \textit{path length} scores during training (Fig. \ref{fig:Results1}.a-b, and Fig. \ref{fig:Results2}.a-d). These improvements were statistically significant. In addition, the average \textit{path length} improvement of the 1Hz group was significantly higher than the improvement of the control group. Because the perturbations directly increases the path length, the improvement of the 1Hz group probably consists of an improvement as a result of learning the task (similar to the control group) and an additional improvement caused by coping with the perturbations. 

The approach metrics showed that the participants in the 1Hz group significantly reduced the number of cuts during training, whereas in the control group, no such change was observed (Fig. \ref{fig:Results2}.e-f). A possible explanation is that participants in the 1Hz group had to time their cuts according to the perturbations, forcing participants who cut quickly to lower the pace. In addition, the participants in the 1Hz group significantly reduced their \textit{perturbation MSE} scores during training. This means that at the end of the training, the 1Hz fluctuations in their movement were smaller than at the beginning of the training. 

\subsection{Q2 -- The effect of the training protocols on post-training trials without perturbations (P1)}
After the training (P1), the groups reached a similar level of performance in \textit{combined error-time} and \textit{path length} (Fig. \ref{fig:Results2}.a,c). There was a small and not significant difference between the average improvement in  \textit{combined error-time} of the two groups (Fig. \ref{fig:Results2}.b). This small difference may stem from the fact that the control group had a slightly higher score in the baseline trials, thereby having more room for improvement during training.
The number of cuts of the 1Hz group was significantly smaller than of the control group, suggesting that after getting used to a lower number of cuts caused by the perturbations, they did not raise it even after the perturbations were removed.

\subsection{Q3 -- The effect of the training protocols on post-training trials with 1Hz perturbations (P2)}
  In all four metrics, the average value of the 1Hz group was slightly lower than the control group, but these differences were not statistically significant. For the number of cuts, the value of the difference ($P2-P1$) of the control group was significantly smaller than the 1Hz group (Fig. \ref{fig:Results2}.f). In Fig. \ref{fig:Results2}.e, we can see that while participants in the 1Hz group kept the same number of cuts for P1 and P2, participants in the control group reduced this number. Apparently, participants who chose an approach of high number of cuts were forced to change it due to the perturbations.

\begin{table*}[h]
	\caption{Statistical Analysis summary.}
 	\label{tab:Results}

\begin{tabular}{|c|c|c|c|c|c|c|c|c|c|}
\hline
\multirow{3}{*}{} &
  \multicolumn{3}{c|}{\textbf{Q1}} &
  \multicolumn{2}{c|}{\textbf{Q2}} &
  \multicolumn{2}{c|}{\textbf{Q3}} &
  \multicolumn{2}{c|}{\textbf{Q 4}} \\ \cline{2-10} 
 &
  \multicolumn{3}{c|}{\begin{tabular}[c]{@{}c@{}}Training start – training end\\    (T\textsubscript{start}-T\textsubscript{end})\end{tabular}} &
  \begin{tabular}[c]{@{}c@{}}Post\\ no pert. \\ (P1)\end{tabular} &
  \begin{tabular}[c]{@{}c@{}}Baseline \\ –post no pert.  \\ (B-P1)\end{tabular} &
  \begin{tabular}[c]{@{}c@{}}Post 1Hz \\ (P2)\end{tabular} &
  \begin{tabular}[c]{@{}c@{}}Post 1Hz \\ – post no pert.  \\ (P2-P1)\end{tabular} &
  \begin{tabular}[c]{@{}c@{}}Post unpred. \\ (P3)\end{tabular} &
  \begin{tabular}[c]{@{}c@{}}Post unpred.\\ – post no pert.\\ (P3-P1)\end{tabular} \\ \cline{2-10} 
 &
  Control &
  1Hz &
  \begin{tabular}[c]{@{}c@{}}1Hz -\\ Control\end{tabular} &
  \begin{tabular}[c]{@{}c@{}}1Hz -\\ Control\end{tabular} &
  \begin{tabular}[c]{@{}c@{}}1Hz -\\ Control\end{tabular} &
  \begin{tabular}[c]{@{}c@{}}1Hz -\\ Control\end{tabular} &
  \begin{tabular}[c]{@{}c@{}}1Hz -\\ Control\end{tabular} &
  \begin{tabular}[c]{@{}c@{}}1Hz -\\ Control\end{tabular} &
  \begin{tabular}[c]{@{}c@{}}1Hz –\\ Control\end{tabular} \\ \hline
\multirow{2}{*}{\textbf{\begin{tabular}[c]{@{}l@{}}Combined \\ error-time\end{tabular}}} &
  \textbf{$\Delta$=0.516} &
  \textbf{$\Delta$=0.962} &
  $\Delta$=0.446 &
  $\Delta$=-0.078 &
  $\Delta$=-0.327 &
  $\Delta$=-0.835 &
  $\Delta$=-0.757 &
  $\Delta$=-0.19 &
  $\Delta$=-0.112 \\
 &
  \textbf{p=0.002} &
  \textbf{p=0.003} &
  p=0.211 &
  p=1 &
  p=0.331 &
  p=0.233 &
  p=0.231 &
  p=1 &
  p=1 \\ \hline
\multirow{2}{*}{\textbf{Path length}} &
  \textbf{$\Delta$=5.65} &
  \textbf{$\Delta$=32.774} &
  \textbf{$\Delta$=27.124} &
  $\Delta$=3.529 &
  $\Delta$=-3.516 &
  $\Delta$=-21.824 &
  $\Delta$=-25.353 &
  $\Delta$=7.85 &
  $\Delta$=4.321 \\
 &
  \textbf{p=0.035} &
  \textbf{p\textless{}0.001} &
  \textbf{p\textless{}0.001} &
  p=0.932 &
  p=0.897 &
  p=0.266 &
  p=0.115 &
  p=1 &
  p=1 \\ \hline
\multirow{2}{*}{\textbf{\begin{tabular}[c]{@{}l@{}}Number \\ of cuts\end{tabular}}} &
  $\Delta$=-0.633 &
  \textbf{$\Delta$=6.133} &
  $\Delta$=6.767 &
  \textbf{$\Delta$=-15.822} &
  $\Delta$=9.711 &
  $\Delta$=-4.644 &
  \textbf{$\Delta$=11.178} &
  $\Delta$=-5.111 &
  \textbf{$\Delta$=10.711} \\
 &
  p=1 &
  \textbf{p\textless{}0.001} &
  p=0.057 &
  \textbf{p=0.015} &
  p=0.23 &
  p=0.478 &
  \textbf{p=0.026} &
  p=0.493 &
  \textbf{p=0.006} \\ \hline
\multirow{2}{*}{\textbf{Pert. MSE}} &
  $\Delta$\textless{}0.001 &
  \textbf{$\Delta$=0.024} &
  \textbf{$\Delta$=0.024} &
  $\Delta$\textless{}0.001 &
  $\Delta$\textless{}0.001 &
  $\Delta$=-0.022 &
  $\Delta$=-0.022 &
  – &
  – \\
 &
  p=0.223 &
  \textbf{p=0.002} &
  \textbf{p\textless{}0.001} &
  p=1 &
  p=0.994 &
  p=0.218 &
  p=0.19 &
  – &
  – \\ \hline
\end{tabular}

$\Delta$ denotes the mean of the individual differences for matched-pair tests, and the difference between the means of the two groups for the other tests. Bold font indicates statistically significant effects ($p < 0.05$).
\end{table*}

\subsection{Q4 -- The effect of the training protocols on post-training trials with unpredictable perturbations (P3)}
The small differences in \textit{combined error-time} and \textit{path length} values between the groups in the P2 stage were further reduced in P3. Looking at Fig. \ref{fig:Results2}.a,c, it seems that the control group improved their performances in P3 relative to P2, while the 1Hz group had no improvement in \textit{combined error-time} and a smaller improvement in \textit{path length}. There was a significant difference between the $P3-P1$ values of the two groups because the control group continue to cut with less cuts than in P1, while the 1Hz group did not change the number of cuts. 

\section{DISCUSSION}
In this letter we presented a new experimental protocol for testing the effect of time-dependent force perturbations on the learning of a RAMIS task. Our aim was to harness motor learning theories to improve the training of RAMIS surgeons. Our analysis revealed several approaches that participants used to overcome the perturbations without impairing their performance when the perturbations were removed. Our results form a basis for further research that could improve the way surgeon acquire RAMIS skills.

So far the effect of assistive and resistive forces on RAMIS training has been investigated \cite{m.m.coadTrainingDivergentConvergent2017,enayatiRoboticAssistanceasNeededEnhanced2018,oquendoRobotAssistedSurgicalTraining2019}; however, to the best of our knowledge, the effect of time-dependent perturbations had not been explored. We found that participants improved their performance during training with 1Hz periodic perturbations. Compared to at the beginning of the training, at the end of the training the perturbations had less impact on their movement. Such a result can be caused by an adaptation in which the motor system learns the force perturbations and applies force in the opposite direction, which reduces the impact of perturbations on the movement. When adaptation occurs, an after-effect can be seen after the perturbations are removed. In our experiment, such after-effect would have been expressed as 1Hz fluctuations in the P1 trials. Because the \textit{perturbation MSE} of the 1Hz group returned to the baseline level as soon as the perturbation was removed, we conclude that adaptation did not occur. This result is consistent with previous studies that showed that participants do not adapt to time varying perturbations \cite{karnielSequenceTimeState2003}.

Although adaptation did not occur, the participants did develop approaches to improve. Previous studies showed that participants stabilize unstable conditions by co-contraction of the muscles and an increase in the impedance of the arm \cite{burdetCentralNervousSystem2001}. Such an increase can be the cause of the decrease in \textit{perturbation MSE} during training in our experiment. Another approach revealed by our analysis is a change in the number of cuts during the training. This result is consistent with \cite{patelUsingIntermittentSynchronization2018}, where a surgeon who performed a pattern-cutting task with a moving platform timed the cuts according to the movement of the platform.

Importantly, we showed that when the perturbations were removed both groups reached a similar performance level. This means that learning how to deal with the perturbations was not at the expense of learning how to perform the task better. This result suggests that trainees can be exposed to challenging conditions during training without impairing their learning. When the participants encountered the 1Hz perturbations after training the advantage of the 1Hz group was not significant. This small advantage was further reduced when participants encountered the unpredictable perturbations. It is possible that after a long training on the task under simple conditions the control group was able to learn quickly how to deal with the perturbations, and that therefore the gap between the groups was small. Additional research is required to examine protocols that will incorporate perturbations in a way that may increase the advantages over learning without perturbations. 

This is the first study of time-dependent perturbations in RAMIS training, and hence we chose simple and not fully realistic perturbations. Now that we have found that training with these perturbations did not impair the learning processes we plan to examine more realistic perturbations.

\section{CONCLUSIONS}
We developed a novel experimental protocol and new analysis tools to examine how time-dependent perturbations affect of the learning of a surgical task. The data collected for this study are available (on request to the corresponding author), and can be used to advance  surgical robotics and motor learning research. We found that participants learned how to overcome these perturbations and that this learning was not at the expense of learning the task. Our results lead the way toward developing training protocols that will incorporate time-dependent perturbations, which could improve the way surgeons acquire RAMIS skills. From the perspective of the motor learning field, this study is an important step toward understanding learning in real life tasks.


\section*{ACKNOWLEDGMENT}
The authors would like to thank Anton Deguet, Simon DiMaio, Eli Peretz, and Gilat Malka for their help with the dVRK integration in our lab, and Alon Lempert and  Noa Yamin for running the experiments.
\bibliographystyle{IEEEtran}
\bibliography{Refs}

\end{document}